%% file: main.tex
\documentclass[10pt,twocolumn,letterpaper]{article}

\usepackage[pagenumbers]{iccv} 


\definecolor{iccvblue}{rgb}{0.21,0.49,0.74}
\usepackage[pagebackref,breaklinks,colorlinks,allcolors=iccvblue]{hyperref}
\usepackage{amssymb} 
\usepackage{array} 

\def\paperID{12356} 
\def\confName{ICCV}
\def\confYear{2025}

\usepackage{color, colortbl}
\definecolor{Gray1}{gray}{0.9}
\definecolor{Gray2}{gray}{1.0}
\usepackage{wrapfig}
\usepackage{graphicx}
\usepackage{multirow}
\usepackage{subcaption}
\usepackage{bbding}

\newcommand{\data}{3DCoMPaT-GrIn}
\newcommand{\task}{PaPGD}

\title{Kestrel: 3D Multimodal LLM for Part-Aware Grounded Description}

\author{
Mahmoud Ahmed$^{*}$,\ \ Junjie Fei$^{*}$,\ \ Jian Ding,\ \ Eslam Mohamed BAKR,\ \ Mohamed Elhoseiny \\
King Abdullah University of Science and Technology (KAUST) \\
\tt\small \{mahmoud.ahmed,junjie.fei,jian.ding,eslam.abdelrahman,mohamed.elhoseiny\}@kaust.edu.sa
}

\begin{document}
\maketitle

\let\thefootnote\relax\footnotetext{$^*$ Equal contribution}

\input{sec/abs.tex}
\input{sec/intro.tex}
\input{sec/related.tex}
\input{sec/dataset.tex}
\input{sec/methods.tex}
\input{sec/experiments.tex}
\input{sec/limitation}
\input{sec/conclusion.tex}
\input{sec/acknowledgements}

{
    \small
    \bibliographystyle{ieeenat_fullname}
    \bibliography{main}
}

\input{sec/supp}

\end{document}

%% file: sec/abs.tex
\begin{abstract}

In this paper, we introduce Part-Aware Point Grounded Description (\task), a challenging task aimed at advancing 3D multimodal learning for fine-grained, part-aware segmentation grounding and detailed explanation of 3D objects.
Existing 3D datasets largely focus on either vision-only part segmentation or vision-language scene segmentation, lacking the fine-grained multimodal segmentation needed for robotic navigation and interaction in real-world environments.
To address this gap, we present the 3DCoMPaT Grounded Instructions (\data) Dataset, a comprehensive resource that pairs rich point cloud descriptions with corresponding part-level segmentation masks. This dataset encompasses extensive samples designed for both PaPGD and fine-grained single-part grounding tasks.
To tackle the inherent challenges of grounding objects and generating grounded descriptions at the part level, we propose Kestrel, a part-aware 3D multimodal large language model that integrates an advanced language model for nuanced language comprehension with multi-level point feature propagation and query refinement mechanism to enhance spatial reasoning at the part level. The extensive experiments demonstrate that Kestrel effectively bridges the gap between part-aware language understanding and 3D segmentation grounding, paving the way for more robust and interpretable 3D object comprehension that meets the demands of real-world robotic applications.
Project page: \url{https://feielysia.github.io/Kestrel.github.io}
%
\end{abstract}

%% file: sec/intro.tex
\section{Introduction}

\input{images/teaser}

Recent progress in 3D data collection have revolutionized the way we represent scenes and objects. While global scene segmentation has made significant strides, achieving compositional 3D grounded reasoning—where models can accurately identify and segment object parts while understanding their material attributes—remains a challenging frontier. For instance, consider the question, \emph{``Which part of a hot teapot can I grasp to hold it?"} An ideal response would be \emph{``the handle, since it is made of wood, offering a secure and safe grip."} This kind of compositional reasoning, which combines material-aware analysis with spatial segmentation, is critical for developing 3D vision-language systems that can interact reliably with real-world objects.

Advancements in large language models (LLMs)~\cite{GPT2,GPT3,instructgpt,touvron2023llama,llama2} and 2D multimodal LLMs (MLLMs)~\cite{dai2024instructblip,li2023blip,zhu2023minigpt,chen2023minigpt,llava,liu2023improved} has led to significant improvements in detailed object-level recognition, pixel-wise segmentation~\cite{lai2023lisa,rasheed2023glamm}, and compositional understanding~\cite{li2023covlm}. These breakthroughs have spurred a growing trend to adapt MLLMs for 3D applications~\cite{xu2023pointllm,guo2023point,3D-LLM,qi2023gpt4point,tang2024minigpt,SegPoint} to bridge the gap between human and machine interpretation of 3D environments. However, a critical limitation persists: \textit{existing 3D MLLMs often fail to capture the fine-grained details of object parts and their material properties, which are essential for precise real-world interaction.} This shortcoming highlights the need for further research to advance 3D MLLMs toward a more detailed, part-aware understanding.

In this work, we introduce the Part-Aware Point Grounded Description (\task) task, where the objective is to generate a detailed, part-aware grounded description of an input point cloud by linking each descriptive phrase to its corresponding segmentation mask (see Fig.~\ref{fig:teaser}). Recognizing that no existing dataset supports training and evaluating this fine-grained 3D vision-language understanding, we propose the 3DCoMPaT Grounded Instructions (\data) dataset. Building on 3DCoMPaT~\cite{li20223d_compat,slim_3dcompatplus_2023}, we enhance point clouds by incorporating multiple rendered views and processing them, along with detailed part annotations and material information, through GPT-4o~\cite{openai2024gpt4o} to produce comprehensive, part-aware grounded descriptions. The resulting dataset comprises 111,514 training samples and 9,449 validation samples for \task, direct segmentation, and reasoning segmentation tasks.
These tasks are essential for evaluating the capability of 3D MLLMs to generate both fine-grained segmentation masks and comprehensive grounded descriptions, a crucial step toward developing the next generation of 3D vision-language systems, with broad applications in robotics, autonomous navigation, and interactive technologies.

To tackle the challenges posed by \task, we propose Kestrel, a novel part-aware 3D MLLM designed to capture the intricate spatial and compositional details required for precise language understanding and segmentation grounding. Kestrel is composed of four key components: a point encoder, an LLM, a point feature propagation module (PFPM), and a 3D segmentation decoder. The point encoder extracts detailed features from the input point cloud without relying on extensive global feature embeddings. The LLM leverages these features to generate part-specific descriptions and segmentation queries. A specialized query refinement mechanism then enhances these queries within segmentation decoder by incorporating different upsampling features from PFPM to ensure that the final segmentation masks accurately reflect the decoded point cloud features.

In summary, our contributions are as follows:
\begin{itemize}
    \item We introduce Part-Aware Point Grounded Description (\task), a novel task that challenges 3D MLLMs to achieve detailed object understanding through material-aware, part-level segmentation and description.
    \item We present 3DCoMPaT Grounded Instructions (\data), a comprehensive dataset designed to support the training and evaluation of fine-grained, part-aware vision-language understanding in 3D field.
    \item We propose Kestrel, a dedicated part-aware 3D MLLM that combines query refinement with multi-level feature decoding to deliver precise language understanding and segmentation grounding, setting a new benchmark in 3D compositional segmentation understanding.
\end{itemize}

%% file: images/teaser.tex
\begin{figure}[t]
    \centering
    \includegraphics[width=1.0\linewidth]{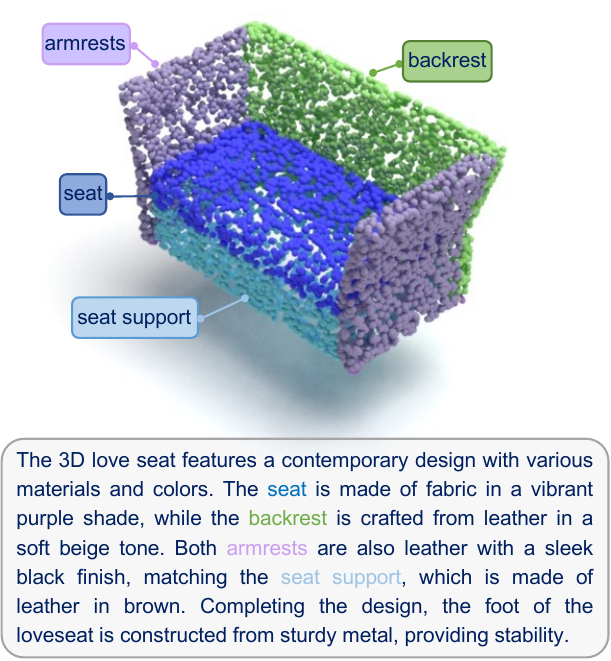}
    \caption{\textbf{Part-Aware Point Grounded Description.}
    Given an input point cloud, the model is tasked with predicting a grounded description - text that provides a detailed interpretation of the 3D object. Each part-level phrase in this generated text (\eg, “back-rest” and “seat support”) is linked to a point-wise segmentation mask, challenging the model’s capability for part-aware language understanding and segmentation grounding (it is worth noting that the colors shown in this figure are not the actual colors of the point cloud but are used to represent the different segmentation masks).
    }
    \label{fig:teaser}
\end{figure}

%% file: sec/related.tex
\section{Related Work}

\noindent\textbf{2D MLLMs.} Recent advancements in LLMs such as GPT~\cite{GPT2,GPT3,instructgpt}, LLaMA~\cite{touvron2023llama, llama2}, Alpaca~\cite{alpaca}, and Vicuna~\cite{vicuna2023}, along with their multimodal extensions~\cite{flamingo,li2023blip,li2022blip,dai2024instructblip,chen2023minigpt,zhu2023minigpt,llava,llava1.5}, have significantly enhanced text generation and multimodal reasoning capabilities.
Key developments in this area include Flamingo~\cite{flamingo}, BLIP~\cite{li2022blip,li2023blip}, MiniGPT-4~\cite{zhu2023minigpt}, InstructBLIP~\cite{dai2024instructblip}, and LLaVA~\cite{llava,llava1.5}, which advance AI's ability to interpret and interact using vision and language.
However, these models are limited in their ability to ground regions in the image. Recent endeavors such as VisionLLM~\cite{wang2024visionllm}, Kosmos-2~\cite{peng2023kosmos}, Shikra~\cite{chen2023shikra}, Qwen-VL~\cite{bai2023qwen}, and MiniGPT-v2~\cite{chen2023minigpt} have sought to empower MLLMs with grounding capabilities, largely by integrating location tokens into the MLLM vocabulary and predicting bounding boxes, a concept inspired by Pix2Seq~\cite{chen2021pix2seq,chen2022unified}.
VisionLLM~\cite{wang2024visionllm} takes a step further by predicting object masks as polygons. However, this approach compromises the efficiency of model inference.
To achieve segmentation grounding more effectively, LISA~\cite{lai2023lisa} and GLaMM~\cite{rasheed2023glamm} incorporate a segmentation model into MLLM to accurately ground object masks alongside their textual descriptions.
On the other hand, there is a progressive shift toward refining MLLMs for more detailed visual reasoning. Nevertheless, most of the current MLLMs still fail to achieve part-level reasoning.
In this work, we focus on empowering MLLMs with 3D part-aware understanding.

\noindent\textbf{3D MLLMs.} Following the advances in 2D MLLMs, recent developments in 3D MLLMs~\cite{3D-LLM,xu2023pointllm,guo2023point,qi2023gpt4point,tang2024minigpt,SegPoint,qi2024shapellm} focus on understanding and locating tasks either in 3D scenes or objects.
Similar to their 2D counterparts, 3D MLLMs leverage a strong feature encoder to map 3D data into the latent space. This is achieved either by utilizing CLIP ViT~\cite{radford2021learning} on 2D views of the 3D scenes or objects, as demonstrated in 3D-LLM~\cite{3D-LLM}, or by aligning a 3D encoder with CLIP using ULIP~\cite{xue2022ulip,xue2023ulip2}, \eg, PointLLM~\cite{xu2023pointllm}.
These works can be categorized into three groups: scene and object understanding models~\cite{3D-LLM}, object-level information models~\cite{xu2023pointllm,guo2023point,qi2023gpt4point}, and models that ground 3D objects within scenes~\cite{SegPoint}.
The most relevant work, PARIS3D~\cite{kareem2024paris3d}, attempts part segmentation by 3D uplifting of 2D segmentations, but this method may suffer from inaccuracies due to domain shifts and only focuses on single-part grounding. We propose an alternative method that solely leverages the 3D modality, preserving critical 3D spatial information. Notably, none of these models offer fine-grained, part-aware segmentation grounding and detailed explanation of 3D objects. We aim to fill this gap in this paper.

\noindent\textbf{3D Vision-Language Datasets.} Early 3D datasets focused primarily on vision-only tasks, including classification~\cite{chang2015shapenet,wu20153d,uy2019revisiting}, detection~\cite{caesar2020nuscenes}, and segmentation~\cite{armeni20163d,dai2017scannet,li20223d_compat,slim_3dcompatplus_2023,mo2019partnet}. With the rise of multimodal learning, these datasets have expanded to support vision-language tasks like 3D captioning~\cite{deitke2023objaverse,deitke2024objaverse,luo2024scalable,sun2023unig3d,chen2021scan2cap} and 3D question answering~\cite{azuma2022scanqa}.
To enhance human-3D MLLM interaction, PointLLM~\cite{xu2023pointllm} and GPT4Point~\cite{qi2023gpt4point} developed annotation pipelines to gather fine-tuning instructions. However, few datasets focus on precise part localization within 3D objects. While datasets like ScanRefer~\cite{chen2020scanrefer} and instruct3D~\cite{SegPoint} can localize 3D objects using natural language, they do not support fine-grained part segmentation grounding.
Though part-aware segmentation datasets exist, such as 3DCoMPaT~\cite{li20223d_compat,slim_3dcompatplus_2023}, and PartNet~\cite{mo2019partnet}, adapting them for part-aware vision-language understanding remains an open challenge.
The 3D RPSeg dataset~\cite{kareem2024paris3d}, built on PartNetE~\cite{liu2023partslip}, introduces the Reasoning 3D Part Segmentation benchmark. However, it only covers one part per sample and is limited by a small number of shapes and parts per category. This highlights a need for a large diverse dataset that accurately provides part-level segmentation masks based on user instructions.
In response, we introduce \data, a new dataset designed to train and evaluate part-aware understanding in 3D MLLMs, providing a foundation for fine-grained segmentation and part-level language generation.

%% file: sec/dataset.tex

\input{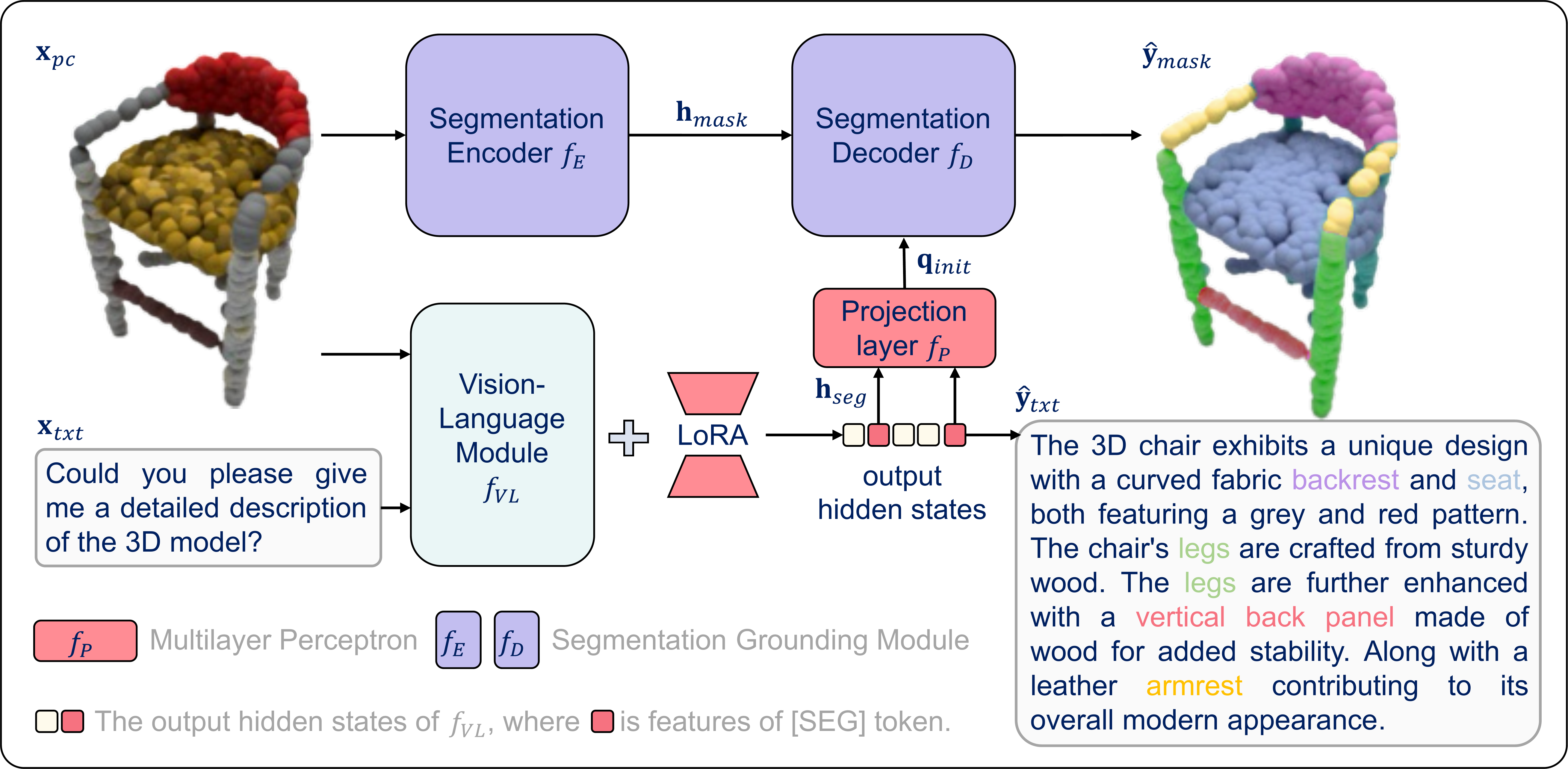}

\section{\data}

To support the training and evaluation of 3D MLLMs in part-aware language understanding and segmentation grounding, we introduce the 3DCoMPaT Grounded Instructions (\data) Dataset by building on 3DCoMPaT~\cite{li20223d_compat,slim_3dcompatplus_2023}, a compositional 3D dataset that includes two types of part-level annotations.
Part annotations $\mathbf{p} = \{p_{1}, p_{2}, \dots, p_{n}\}$, which identify the constituent components of objects, such as a “handle,”
Material annotations $\mathbf{m} = \{m_{1}, m_{2}, \dots, m_{k}\}$ that specify the materials comprising an object, such as “metal.”
Each annotation corresponds to a segmentation mask within the point cloud.

\noindent\textbf{Part-Aware Point Grounded Description (PaPGD)}
The \data~provides a comprehensive grounded description for each unique point cloud. As shown in Fig~\ref{fig:teaser}, it links every part-level reference in the text to a corresponding segmentation mask within the point cloud.
Specifically, for each point cloud, we generate grounded descriptions using fine-grained part annotations $\textbf{p}$ and material annotations $\textbf{m}$ from 3DCoMPaT to establish a mapping between each part and its corresponding material.
To enhance focus on prominent features, we limit the part annotations per point cloud to a maximum of six, based on the largest visible areas in the 2D rendered images.
This mapping, along with four rendered views of the point cloud, is provided to GPT-4o to generate a detailed text that includes an overview of the shape, parts, material composition, colors, and potential functionality.
We create a set of 30 user instructions, guiding the model to perform part-aware point grounded description.
In total, we collect 80,760 point cloud-grounded description pairs for training and 6,770 for validation.

\noindent\textbf{Direct Segmentation.}
Additionally, we incorporate direct single-part grounding data to help the model segment user-specified parts of the point cloud.
For the training set, we sample one part annotation per point cloud, creating 80,760 unique samples. We then selected a 10\% subset covering all shapes, resulting in 8,076 direct segmentation samples with 15 prompt templates. Using the same method, we collect 677 validation samples for direct segmentation. 

\noindent\textbf{Reasoning Segmentation.}
To enable better model generalizability and understanding of the reasoning segmentation task, We also introduce a more challenging subset for single-part grounding, consisting of 22,678 training samples and 2,002 validation samples, covering 125 parts. To collect this subset, we use GPT-4o to create a unique prompt and response for each part in the shapes and then we manually filter out categories which have identical functionalities per shape by keeping only one of them.

To evaluate the quality of the validation set of the grounded descriptions generated by GPT-4o, we provide our validation data to human annotators to verify the quality of each description by checking the existence of all ground truth parts, their associated materials, and representative colors, using 8 rendered view images for each point cloud. We do this iteratively to ensure a 100\% accurate validation set for this task. The two single-part grounding tasks are generated based on templates and 3DCoMPaT's metadata.
These three subsets are used to establish our proposed \textit{\data}. Overall, we collect a total of 111,514 training samples and 9,449 validation samples.

%% file: images/arch.tex
\begin{figure*}[t]
    \centering
    \includegraphics[width=1.0\linewidth]{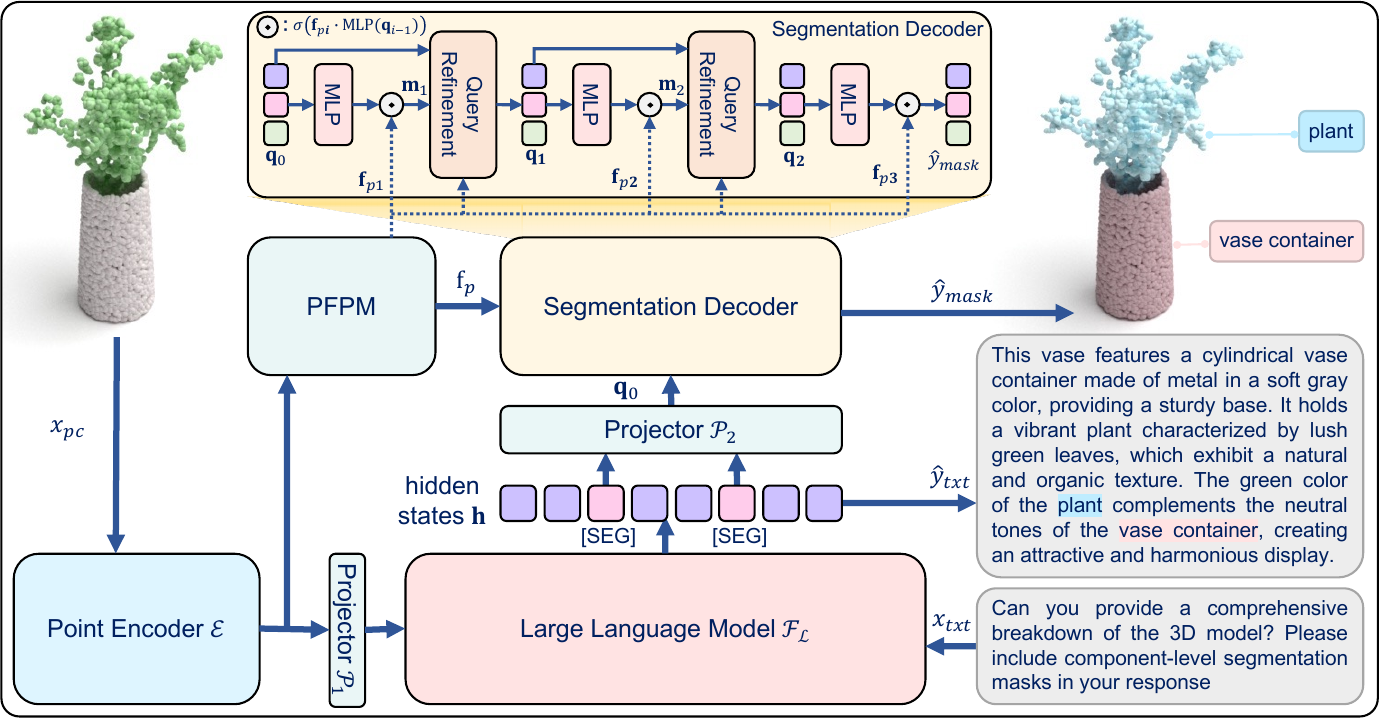}
    \caption{\textbf{Kestrel: A Part-Aware Point Grounding MLLM.}
    The Kestrel model incorporates a point encoder and an LLM to construct a 3D MLLM, designed to generate detailed descriptions based on the input point cloud and text.
    The 3D Segmentation Decoder extracts the output embedding of the \texttt{[SEG]} token from the output hidden states of the 3D MLLM. After projecting these \texttt{[SEG]} embeddings, the 3D SGM uses them as initial queries $\mathbf{q}_0$.
    The point feature propagation module (PFPM) encodes multi-level point features $\mathbf{f}_{p}$.
    Then, the segmentation decoder takes $\mathbf{q}_0$ and $\mathbf{f}_{p}$ as input to generate the point-wise segmentation masks using a query refinement mechanism.
    }
    \label{fig:kestrel}
\end{figure*}


%% file: sec/methods.tex
\def\pointencoder{$\mathcal{E}$}
\def\mllm{$\mathcal{F_{VL}}$}
\def\llm{$\mathcal{F_L}$}
\def\decoder{$\mathcal{D}$}

\section{Method}

Despite the significant advancements in tasks such as 3D captioning, 3D question answering, and 3D object grounding, current 3D MLLMs fall short in accurately predicting point-wise segmentation masks at the part level.
To bridge this gap, we propose Kestrel, which combines a 3D MLLM with a query refinement mechanism to enable fine-grained part segmentation along with detailed textual description.
In Sec.~\ref{methods:kestrel}, we formally introduce Kestrel as a part-aware point grounding 3D MLLM, followed by a detailed explanation of our training objective in Sec.~\ref{methods:objective}.

\subsection{Kestrel}
\label{methods:kestrel}

\noindent\textbf{Architecture Overview.} As shown in Figure~\ref{fig:kestrel}, Kestrel is composed of a point encoder, an LLM, a point feature propagation module (PFPM), and a segmentation decoder.
The point encoder and LLM take a point-aware instruction and point cloud as input, generating a detailed part-level description of the point cloud.
The segmentation decoder leverages a query refinement mechanism to add more fine-grained signals to the queries by integrating different-level upsampling features from PFPM to obtain better part masks.

\input{images/exp1}

\noindent\textbf{3D MLLM.} The 3D MLLM (\mllm) leverages a point encoder (\pointencoder) to map input point cloud into the LLM space(\llm). We introduce projector $\mathcal{P}_1$ to align the latent space of language and 3D vision.
To generate part-aware grounded captions, three specialized tokens \texttt{<p>}, \texttt{</p>}, and the segmentation token \texttt{[SEG]} are incorporated into the 3D MLLM's vocabulary
inspired by LISA~\cite{lai2023lisa} and GLaMM~\cite{rasheed2023glamm}.
The 3D MLLM processes point cloud inputs, $x_{pc}$, and text inputs, $x_{txt}$, to generate a detailed caption, $\hat{y}_{txt}$, which includes fine-grained descriptions of object parts. This process is formulated as
\begin{equation}
\hat{y}_{txt} = \mathcal{F_{VL}}(x_{pc}, x_{txt}) = \mathcal{F_L}(\mathcal{P}_1(\mathcal{E}(x_{pc})), x_{txt}).
\label{eq:mllm} 
\end{equation}
When generating grounded captions, \texttt{[SEG]} tokens are included in the predictions. The hidden states, $\mathbf{h}_{seg}$, associated with these tokens, are projected into the 3D segmentation decoder’s latent space using another projector, $\mathcal{P}_2$, to produce initial segmentation queries, $\mathbf{q}_{0}$, as follows
\begin{equation}
    \mathbf{q}_{0} = \mathcal{P}_2(\mathbf{h}_{seg}).
\label{eq:proj}
\end{equation}

\noindent\textbf{Query Refinement Mechanism.} In \pointencoder, the point cloud is downsampled from $N \times 6$ to $\hat{N} \times E$, where $N$ is the original number of points, $\hat{N}$ is the reduced point set, and $E$ is the feature dimension.
We then incorporate the point feature propagation module (PFPM)~\cite{qi2017pointnetplusplus} to progressively upsample the features from \pointencoder\, to intermediate features $\mathbf{f}_{p1}$, $\mathbf{f}_{p2}$, and finally $\mathbf{f}_{p3}$, with sizes $N_1$, $N_2$, and $N$, respectively.
Each upsampled feature is combined with intermediate segmentation decoder queries, $\mathbf{q}_{i} (i \in \{1, 2\})$, which will be projected through an MLP and then combined by a dot product to produce intermediate masks $\mathbf{m}_i$, which serve as attention masks for the following query refinement module. The mask generation process is defined as
\begin{equation}
    \mathbf{m}_i = \mathbb{I}(\sigma(\mathbf{f}_{pi} \cdot \text{MLP}(\mathbf{q}_{i-1})^T) > 0.5),
\end{equation}
where $\sigma$ denotes the sigmoid function, and $\mathbb{I}(* > 0.5)$ outputs 1 if $*$ is greater than 0.5, otherwise 0.

As shown at the top of Fig.~\ref{fig:kestrel}, the query refinement module takes the initial queries $\mathbf{q}_{0}$ obtained from 3D MLLM and enhances them using the intermediate masks $\mathbf{m}_i$. The refinement process for each feature level is formulate as
\begin{equation}
    \mathbf{q_i} = \text{MLP}(\text{SelfAtt}(\text{CrossAtt}(\mathbf{q_{i-1}}, \mathbf{m}_i, \mathbf{f}_{pi}))),
\end{equation}
where $i \in \{1, 2\}$. After the second refinement stage, the refined queries, together with $\mathbf{f}_{p3}$, are used to generate the final segmentation mask prediction $\hat{y}_{mask}$.
This masked attention process allows each query to only attend to its relevant part features at each upsampled level, leading to accurate segmentation masks.

\subsection{Training Objective}
\label{methods:objective}

Our goal is to train an end-to-end MLLM that can generate detailed descriptions while simultaneously predicting point-wise segmentation masks at the part level.
To achieve this, we utilize an auto-regressive cross-entropy loss $L_{CE}$ for text generation, along with binary cross-entropy loss $L_{BCE}$ and Dice loss $L_{Dice}$~\cite{deng2018learning} for segmentation mask prediction. The overall training loss for Kestrel is defined as
\begin{equation}
    L_{lang} = w_{CE} \cdot L_{CE}(\hat{y}_{txt}, y_{txt})
\end{equation}
\begin{equation}
\begin{split}
    L_{mask} = w_{BCE} \cdot L_{BCE}(\hat{y}_{mask}, y_{mask}) \\
    + w_{Dice} \cdot L_{Dice}(\hat{y}_{mask}, y_{mask})
\label{eq:objective} 
\end{split}
\end{equation}
\begin{equation}
    L = L_{lang} + L_{mask}.
\end{equation}
In this formulation, $w_{CE}$, $w_{BCE}$, and $w_{Dice}$ are the weights assigned to each loss term. Following settings in LISA and GLaMM, we set these weights to 1.0, 2.0, and 0.5, respectively, as default settings in our experiments.

%% file: images/exp1.tex
\begin{figure*}[t]
    \centering
    \includegraphics[width=1.0\linewidth]{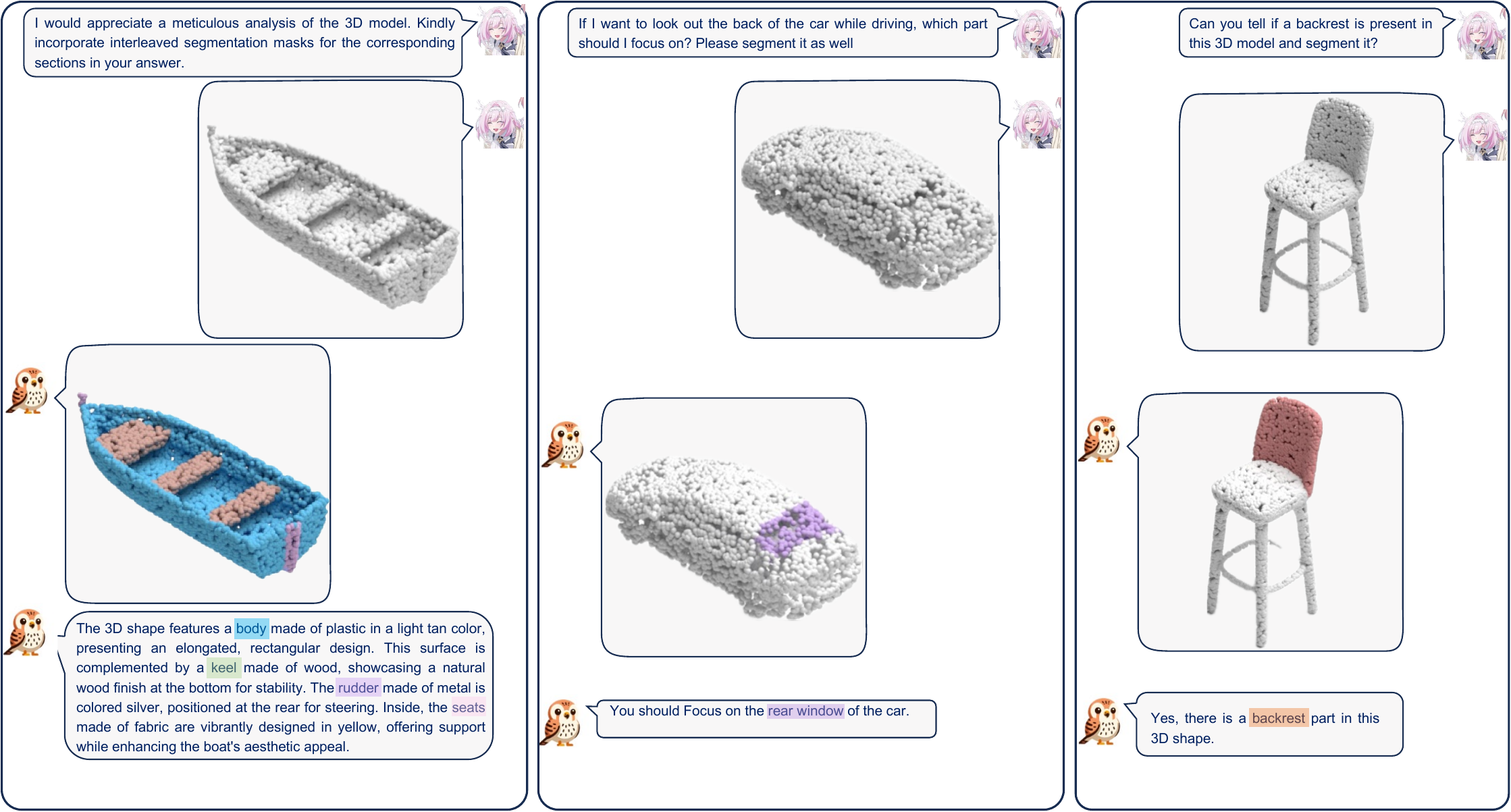}
    \caption{\textbf{Qualitative results of Kestrel on Part-Aware Point Grounded Description, Reasoning and Direct Segmentation.} The results show that Kestrel is capable of detailed 3D object understanding, providing comprehensive description and accurate part-level grounding.
    }
    \label{fig:exp1}
\end{figure*}

%% file: sec/experiments.tex
\section{Experiments}
\input{table/grounded_captions}
\input{table/gpt_eval}

We conduct extensive experiments to evaluate the capability of Kestrel in part-aware vision-language understanding on \data.
In Sec.~\ref{exp:grounded_caption}, part-aware point grounded description evaluates the ability of Kestrel for comprehensive 3D object interpretation in terms of both language understanding and segmentation grounding.
We introduce a new evaluation to capture the 3D Composition-Aware Language Comprehension (3D-CALC).
Sec.~\ref{sec:direct_seg} investigates the performance of Kestrel in single-part grounding from both direct segmentation (\data~and PartNet-Mobility~\cite{Xiang_2020_SAPIEN}) and reasoning segmentation perspectives (\data~and RPSeg3D~\cite{kareem2024paris3d}).
We conduct ablation experiments on our training strategy and Kestrel to explore the effects of design choices, as detailed in Sec.~\ref{exp:ablation}.
In Sec.~\ref{exp:app}, we showcase the robustness and potential applications of Kestrel when the point cloud distribution deviates from the training data, including scenarios where the point clouds are collected from noisy real-world environments.

\noindent\textbf{Implementation Details.} We use Vicuna 7B checkpoint as our LLM along with Uni3d-g\cite{zhou2023uni3d} for point cloud encoding. We pretrain Kestrel on PointLLM's dataset\cite{xu2023pointllm} and \data's point cloud - text pairs (without masks) to capture a better understanding of part-level information in the shapes. The feature propagation and segmentation are as described in Sec.~\ref{methods:kestrel} with an inner dimension of 384.
The projector is implemented as a multi-layer perceptron. We employ LoRA~\cite{hu2021lora} for efficient fine-tuning with the rank set to 8 by default.
Additionally, we utilize AdamW~\cite{loshchilov2017decoupled} optimizer with the learning rate and weight decay set to 0.00009 and 0.0 respectively for the LLM and 0.0002, 0.0 respectively for the 3D SGM. We adopt a cosine learning rate scheduler, with the warmup iteration ratio set to 0.03. All attentions in the LLM are replaced by flash-attention~\cite{dao2022flashattention} during training. The training is done on 4 A100 GPUs for 5 epochs for all experiments with a batch size of 16.

\noindent\textbf{Baselines and Metrics.} As we are the first to develop a 3D MLLM capable of part-aware point grounded description, there are no existing results for direct comparison. The closest methods to our task are SegPoint~\cite{SegPoint}, PARIS3D~\cite{kareem2024paris3d}, and ShapeLLM~\cite{qi2024shapellm}. SegPoint is able to ground objects in scenes, but their code is not yet available, therefore, we implemented their model based on the details in its paper.
ShapeLLM is able to ground object parts using a bounding box but is unable to ground multiple parts due to the number of tokens produced exceeding the maximum context length of the model.
We further introduce an additional baseline, a 3D variation of GLaMM\cite{rasheed2023glamm} by using our LLM with Mask3D~\cite{Schult23ICRA} pretrained on ScanNet instead of SAM~\cite{kirillov2023segany}. We denote this model as Mask3D Baseline. We pass the projected segmentation queries from the MLLM to the Mask3D segmentation decoder as the positional query input for the model, instead of its default initialization method, to guide the segmentation process with the knowledge obtained from the MLLM. Further implementation for this model can be found in the supplementary.

For assessing part-aware language understanding, we employ traditional metrics that measure n-grams overlap and their variations, \ie, BLEU-1~\cite{papineni2002bleu}, ROUGE-L~\cite{lin2004rouge}.
Besides, Sentence-BERT~\cite{reimers2019sentence} and SimCSE~\cite{gao2021simcse} are used to assess sentence similarity at the embedding level.
In addition, we propose a new evaluation for the 3D Composition-Aware Language Comprehension (3D-CALC) capabilities of 3D MLLMs. 
We evaluate the part-aware segmentation grounding using mIoU, AP50, and Recall.
The difference is that we adapt these 2D segmentation grounding metrics from GLaMM\cite{rasheed2023glamm} to our 3D application.

\subsection{Part-Aware Point Grounded Description}
\label{exp:grounded_caption}

We fine-tune Kestrel on the proposed \data~for 5 epochs, using approximately 111K point cloud grounded description pairs.
The results are displayed in Tab.~\ref{tab:grounded_caption}.
Kestrel demonstrates superior part-aware vision-language understanding compared to other 3D baselines (\ie SegPoint\cite{SegPoint}, PARIS3D\cite{kareem2024paris3d}, and Mask3D Baseline) across both language comprehension and segmentation grounding metrics.
In particular, Kestrel achieves the highest mIoU (86.70), AP50 (54.42), and Recall (67.37) among all methods, showcasing its effectiveness in part-aware segmentation grounding.
Kestrel increases the mIoU over the strongest baseline by 3.19\%, highlighting the strength of the query refinement mechanism for fine-grained segmentation.

When compared to 3D MLLMs that can not perform segmentation grounding such as PointLLM-7B\cite{xu2023pointllm} and ShapeLLM-7B\cite{qi2024shapellm}, Kestrel delivers comparable performance on traditional language metrics while achieving the highest score on our proposed GPT evaluation, 3D CALC,
which demonstrates Kestrel’s ability to balance both linguistic and spatial understanding within the 3D domain.

\noindent\textbf{3D Composition-Aware Language Comprehension.} The part-aware language comprehension is evaluated using GPT-4o across three core aspects: (1) Part Accuracy, (2) Material Accuracy, and (3) Composition Accuracy, along with an overall score that averages these categories.
Unlike traditional metrics, which fail to capture the fine-grained alignment between the generated caption and the ground truth, our evaluation measures accuracy at a more detailed compositional level.
For each predicted caption, we compute a score from 0 to 100 for its alignment with the ground truth across individual parts, materials, and part-material pairs. Each score is computed as an average of 5 evaluations with GPT-4o to account for the variations in the score response. The overall score represents the average of these detailed scores, offering a more holistic evaluation of compositional language comprehension. 
The results in Tab.~\ref{tab:gpt_eval} demonstrate that our proposed model, Kestrel, significantly outperforms other models on all aspects of 3D composition-aware language comprehension. Specifically, Kestrel achieves the highest scores across all metrics.
In conclusion, Kestrel effectively bridges the gap between part-aware language comprehension and segmentation grounding in 3D modalities. Its performance validates the efficacy of our proposed architecture.
Qualitative experiments in Fig.~\ref{fig:exp1} and Appendix~\ref{supp:qualtitative} also showcase the Kestrel's strength in predicting grounded descriptions.

\input{table/grounding_point}
\subsection{Single-Part Segmentation Grounding}
\label{sec:direct_seg}

\noindent\textbf{Direct Segmentation}
Tab.~\ref{tab:grounding_point} presents the results for the \textit{Direct Segmentation} task. Kestrel attains the highest scores across all metrics, showcasing its robust ability to interpret instructions and generate accurate, part-level segmentation masks. Notably, it surpasses SegPoint~\cite{SegPoint} by 8.44\% in mIoU and 8.18\% in AP50. While the Mask3D Baseline demonstrates improvements over ShapeLLM~\cite{qi2024shapellm}, it still underperforms both SegPoint and Kestrel, underscoring the importance of integrating a pre-trained multimodal point encoder. PARIS3D~\cite{kareem2024paris3d} exhibits the lowest performance, primarily because our benchmark includes parts not visible in the rendered views, emphasizing the necessity of true 3D data for fine-grained part segmentation grounding.
To further validate Kestrel's effectiveness, we report its performance on PartNet-Mobility~\cite{Xiang_2020_SAPIEN}. The results in Appendix~\ref{supp:morebenchmarks} demonstrate that Kestrel remains a competitive model.

\input{table/rpseg_eval}

\noindent\textbf{Reasoning Segmentation}
\label{sec:reason_seg}
For the \textit{Reasoning Segmentation} task, Tab.~\ref{tab:grounding_point} shows that Kestrel again excels, outperforming SegPoint by 6.60\% in mIoU and 8.17\% in AP50. This highlights the effectiveness of Kestrel’s query refinement and feature decoding. By contrast, PARIS3D performs poorly, likely because its dataset focuses on parts not visible in the rendered point cloud.
We also report the results on RPSeg3D~\cite{kareem2024paris3d} dataset. To make a fair comparison, we fine-tune our proposed Kestrel on RPSeg dataset for one epoch and evaluate it against PARIS3D on their PartNetE-based benchmark. This step is necessary to address the significant color shift between PartNetE and \data. As shown in Tab.~\ref{tab:rpseg}, our model outperforms PARIS3D on the RPSeg3D reasoning segmentation benchmark.

Overall, Kestrel’s strong performance in both direct and reasoning segmentation confirms its ability to produce reliable, part-aware segmentation masks based on user instructions.
Refer to Fig.\ref{fig:exp1} and Appendix~\ref{supp:qualtitative} for comprehensive visualization results of Kestrel on single-part grounding task.

\input{images/generalization}

\subsection{Ablation Studies}
\label{exp:ablation}

\noindent\textbf{Query Refinement Level.} Tab.~\ref{reb:ablation1} examines the effect of varying the number of query refinement stages on the performance of our approach. We maintain the same number of upsampling decoding levels as PointNet++\cite{qi2017pointnetplusplus} throughout the ablations. Without refinement levels, we observe moderate results across all tasks. Introducing a single refinement level provides a clear improvement, while employing two refinement levels achieves the highest mIoU scores for \textit{Grounded Desc.}, \textit{Direct Segmentation}, and \textit{Reasoning Segmentation}. This progression highlights the importance of incrementally refining queries to capture fine-grained part details and accurately map them to textual descriptions.

\noindent\textbf{More Ablations.} To further investigate the impact of different model architectures, we conduct ablation studies on various point encoders and projectors. We also experiment with different amounts and subsets of training data. Refer to Appendix~\ref{supp:moreablation} for more details on these ablation studies.

\input{table/query_ablations}

\subsection{Application}
\label{exp:app}

\textbf{Domain Shift Generalizability.}
We evaluate Kestrel's ability to generalize to new domains by testing it on Objaverse using a checkpoint trained only on \data. As shown in the first two images of Fig.~\ref{fig:generalize}, Kestrel accurately segments parts such as the pulling lever and wheel, indicating robust performance under different data distributions.

\noindent\textbf{Single-Part Grounding Beyond Single Objects.}
Because all point clouds in \data~consist of a single 3D object, we further examine Kestrel’s capacity to handle multiple objects. Specifically, we prompt Kestrel to segment a particular part across multiple 3D objects. As depicted in Fig.~\ref{fig:generalize}, Kestrel can correctly segment parts that appear in both 3D objects or only in one, demonstrating the model’s flexible generalization capabilities in unseen settings.
%

\noindent\textbf{Real-World Demos.} To assess the robustness of Kestrel on real-world data, we evaluate and visualize the model-generated grounded descriptions of objects from the ScanObjectNN~\cite{uy2019revisiting} in Fig.~\ref{fig:realworld}. Kestrel demonstrates the capability to handle noisy and incomplete real-world inputs.

%% file: table/grounded_captions.tex
\begin{table*}[t]
    \centering
    \resizebox{1.0\linewidth}{!}{
        \begin{tabular}{c|ccccc|ccc}
        \toprule
        \multirow{2}{*}{\textbf{Model}} & \multicolumn{5}{c|}{\textbf{Language Understanding}} & \multicolumn{3}{c}{\textbf{Segmentation Grounding}} \\
        ~ & BLEU-4 & METEOR & Sentence-BERT & SimCSE &  GPT EVAL & mIoU & AP50 & Recall \\
        \midrule
        PointLLM-7B\cite{xu2023pointllm} & 6.25 & 29.51 & 83.11 & 84.68 & 48.90 & \textcolor{red}{\XSolidBrush} & \textcolor{red}{\XSolidBrush} & \textcolor{red}{\XSolidBrush} \\
        ShapeLLM-7B\cite{qi2024shapellm}  & 8.01 & 30.05 &\textbf{84.74} & \textbf{85.38} & 47.90 & \textcolor{red}{\XSolidBrush} & \textcolor{red}{\XSolidBrush} & \textcolor{red}{\XSolidBrush} \\
        \midrule
        PARIS3D\cite{kareem2024paris3d} & 8.45 & 31.62 & 82.08 & 83.86 & 49.50 & 47.26 & 11.9 & 38.05 \\
        Mask3D Baseline \textit{(ours)}& 3.10 & 22.73 & 73.71 & 76.03 & 32.80 & 57.90 & 20.00 & 35.20 \\
        SegPoint*\cite{SegPoint}  & 7.24 & 27.62 & 78.50 & 81.47 & 42.80 & 83.51 & 47.16 & 60.66 \\
        \rowcolor{Gray1}
        Kestrel \textit{(ours)}& \textbf{8.55} & \textbf{32.88} & 82.19 & 84.13 & \textbf{50.10} & \textbf{86.70} & \textbf{51.27} & \textbf{67.37} \\
        \bottomrule
        \end{tabular}
    }
    \caption{\textbf{Part-Aware Point Grounded Description Results.} Comparison of models on language understanding and multi-part grounding. Results marked with \textcolor{red}{\XSolidBrush} indicate metrics for the model cannot be evaluated.
    Models marked with * denote our implementations due to unavailable original code. Results show Kestrel excels in part-aware grounding while preserving strong language comprehension.}

    \label{tab:grounded_caption}

\end{table*}

%% file: table/gpt_eval.tex
\begin{table}[h]
    \centering
    \resizebox{\linewidth}{!}{%
        \begin{tabular}{c|cccc}
        \toprule
        \textbf{Model} & \textbf{Part} & \textbf{Material} & \textbf{Composition} & \textbf{Overall}\\
        \midrule
        PointLLM\cite{xu2023pointllm} & 61.50 & 45.60 & 39.60 & 48.90 \\
        ShapeLLM\cite{qi2024shapellm} & 60.80 & 45.70 & 37.20 & 47.90 \\
        \midrule
        PARIS3D\cite{kareem2024paris3d} & 62.76 & 45.90 &	40.87 & 49.50 \\ 
        Mask3D Baseline \textit{(ours)}& 46.60 & 28.00 & 24.00 & 32.80 \\
        SegPoint*\cite{SegPoint}       & 56.10 & 38.20 & 34.20 & 42.80 \\
        \rowcolor{Gray1}
        Kestrel \textit{(ours)}& \textbf{62.90} & \textbf{46.80} & \textbf{40.60} & \textbf{50.10}\\
        \bottomrule
        \end{tabular}
    }
    \caption{\textbf{3D Composition-Aware Language Comprehension (3D-CALC).}
    Part, material, and composition understanding evaluated based on accuracy on \data.}
    \label{tab:gpt_eval}
\end{table}

%% file: table/grounding_point.tex
\begin{table}[b]
    \centering
    \resizebox{1.0\linewidth}{!}{%
        \begin{tabular}{c|ccc|ccc}
        \toprule
        \multirow{2}{*}{\textbf{Model}} & \multicolumn{3}{c|}{\textbf{Direct Segmentation}} & \multicolumn{3}{c}{\textbf{Reasoning Segmentation}} \\
        & mIoU & AP50 & Recall & mIoU & AP50 & Recall \\
        \midrule
        ShapeLLM\cite{qi2024shapellm} & 16.4 & 10.51 & 11.62  & 12.54 & 8.34 & 8.15  \\
        \midrule
        PARIS3D\cite{kareem2024paris3d} & 37.99 & 17.7	& 38.40 & 46.00 & 26.30 & 48.80 \\
        Mask3D Baseline \textit{(ours)} & 46.00 & 26.30	& 48.80 & 31.45 & 12.40 & 32.74 \\
        SegPoint*\cite{SegPoint} & 70.34 & 55.73 & 71.60 & 65.20 & 42.88 & 67.80\\
        \rowcolor{Gray1}
            Kestrel \textit{(ours)} & \textbf{78.78} & \textbf{63.91} & \textbf{81.40}   & \textbf{71.80} & \textbf{51.05} & \textbf{73.20} \\
        \bottomrule
        \end{tabular}
        }
    \caption{\textbf{Single-Part Grounding Results.} Evaluates model performance on implicit grounding and grounded reasoning tasks.
    }
    \label{tab:grounding_point}
\end{table}

        



%% file: table/rpseg_eval.tex
\begin{table}[t]
    \centering
    \resizebox{0.6\linewidth}{!}{%
        \begin{tabular}{c|cc}
        \toprule
        \textbf{Method} & PARIS3D~\cite{kareem2024paris3d} & Kestrel \\
        \midrule
        mIoU  & 56.77  & 72.54 \\
        \bottomrule
        \end{tabular}
    }
    \caption{\textbf{Perfomance on RPSeg3D}}
    \label{tab:rpseg}
\end{table}

%% file: images/generalization.tex
\begin{figure*}[t]
    \centering
    \includegraphics[width=1.0\linewidth]{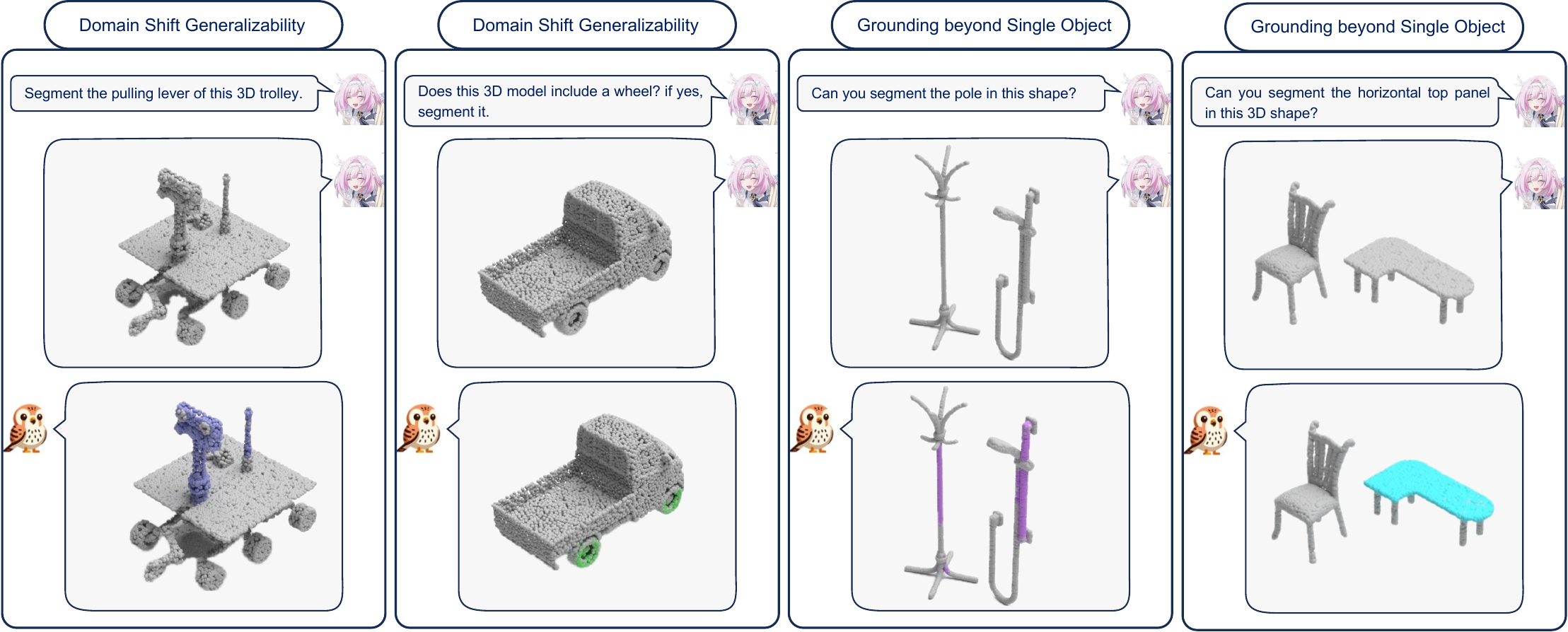}
    \caption{\textbf{Out-of-Domain Generalization.} Kestrel demonstrates robustness when there is a domain shift from \data~to Objaverse, as well as the input distribution offsets from 3D single-object training to 3D multi-object testing.
    }
    \label{fig:generalize}
\end{figure*}
\begin{figure}[t]
    \centering
    \includegraphics[width=1.0\linewidth]{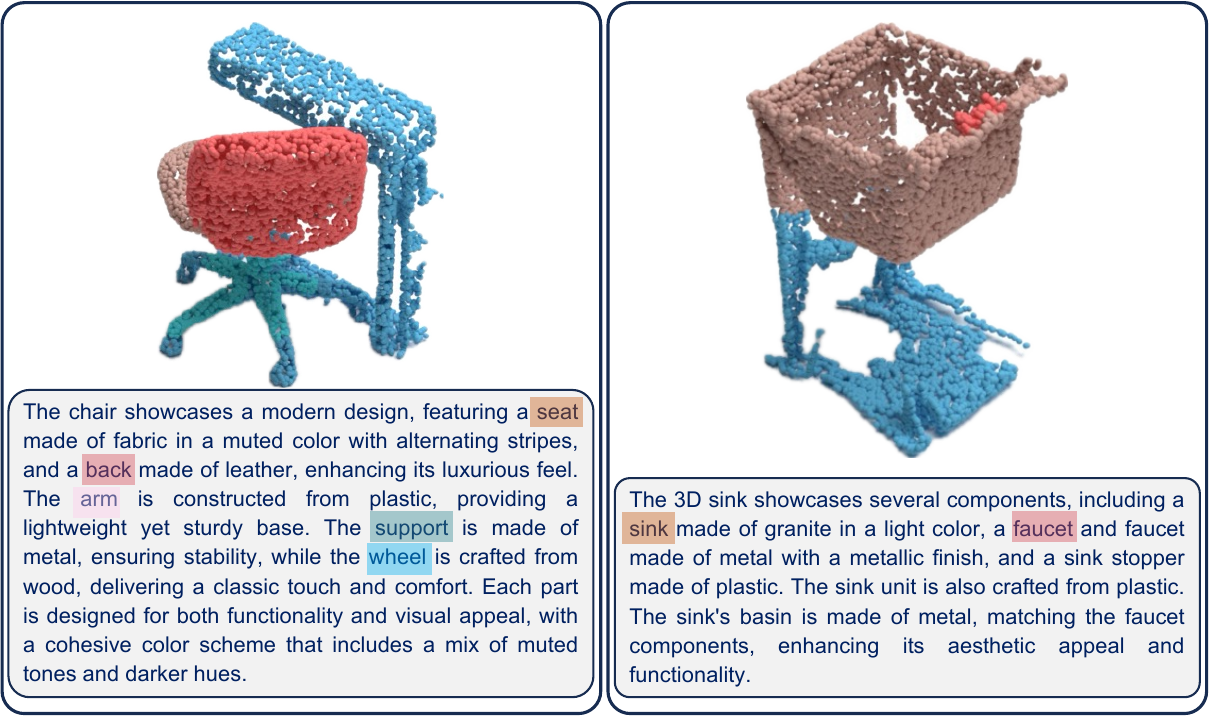}
    \caption{\textbf{Real-Word Demos.} Kestrel shows a certain degree of robustness to noisy and incomplete real-world inputs.}
    \label{fig:realworld}
\end{figure}

%% file: table/query_ablations.tex
\begin{table}[t]
    \centering
    \resizebox{\linewidth}{!}{%
        \begin{tabular}{c|c|c|c}
        \toprule
        \textbf{\# Refinement Levels} & \textbf{Grounded Desc.} & \textbf{Direct Segmentation} & \textbf{Reasoning Segmentation}\\
        \midrule
        0 & 68.44 & 71.60 & 64.53 \\
        1 & 79.43 & 76.55 & 68.90\\
        \rowcolor{Gray1}
        2 & \textbf{86.70} &  \textbf{80.70} & \textbf{71.80} \\
        \bottomrule
        \end{tabular}
    }
    \vspace{-7pt}
    \caption{\textbf{Ablation on the query refinement levels.} Evaluates the effect of changing the number of query refinement stages on the mIoU performance of each task}
    \label{reb:ablation1}
\end{table}

%% file: sec/limitation.tex
%

%% file: sec/conclusion.tex
\section{Conclusion}
In summary, we introduce Part-Aware Point Grounded Description, a new task designed to challenge and advance the capabilities of 3D MLLMs in achieving precise part-level segmentation grounding and detailed object comprehension.
To support this task, we propose the (\data), which provides part-level grounded descriptions for over 120K samples, for the training and evaluation of 3D MLLMs in this part-aware vision-language understanding.
We further present Kestrel, a part-aware point grounding 3D MLLM that leverages a query refinement mechanism to effectively bridge the gap between part-level language understanding and segmentation grounding.
Extensive experiments on PaPGD, Direct Segmentation, and Reasoning Segmentation demonstrate that Kestrel surpass existing methods on the segmentation grounding metrics, as well as achieving the highest scores in GPT-based evaluations for part-aware language comprehension.
Our work establishes a robust benchmark for part-aware 3D vision-language understanding, paving the way for future research in fine-grained 3D object interaction and grounding. By highlighting the potential of part-level understanding, \data~and Kestrel provide a solid foundation for developing AI systems capable of interacting with complex 3D environments in a more human-like manner.

%% file: sec/acknowledgements.tex
\section{Acknowledgements}

The research reported in this publication was supported by funding from King Abdullah University of Science and Technology (KAUST) - Center of Excellence for Generative AI, under award number 5940.

%% file: sec/supp.tex
\clearpage
\setcounter{page}{1}
\setcounter{section}{0}
\renewcommand{\thesection}{\Alph{section}}
\maketitlesupplementary

\input{images/supp_grounded_description}
\input{images/supp_single_part}

\input{sec/supp_baselines_arch}

\input{sec/supp_more_ablate}
\input{sec/supp_more_benchmark}

\input{sec/supp_qualt}

\input{images/supp_gt}
\input{images/website} 

\input{sec/supp_data}

\input{table/gpt_prompt}
\input{table/gpt_prompt2}
\input{table/prompts_multi}
\input{table/prompts_single}
\input{table/template_single}

%% file: images/supp_grounded_description.tex
\begin{figure*}[t]
    \centering
    \includegraphics[width=1.0\linewidth]{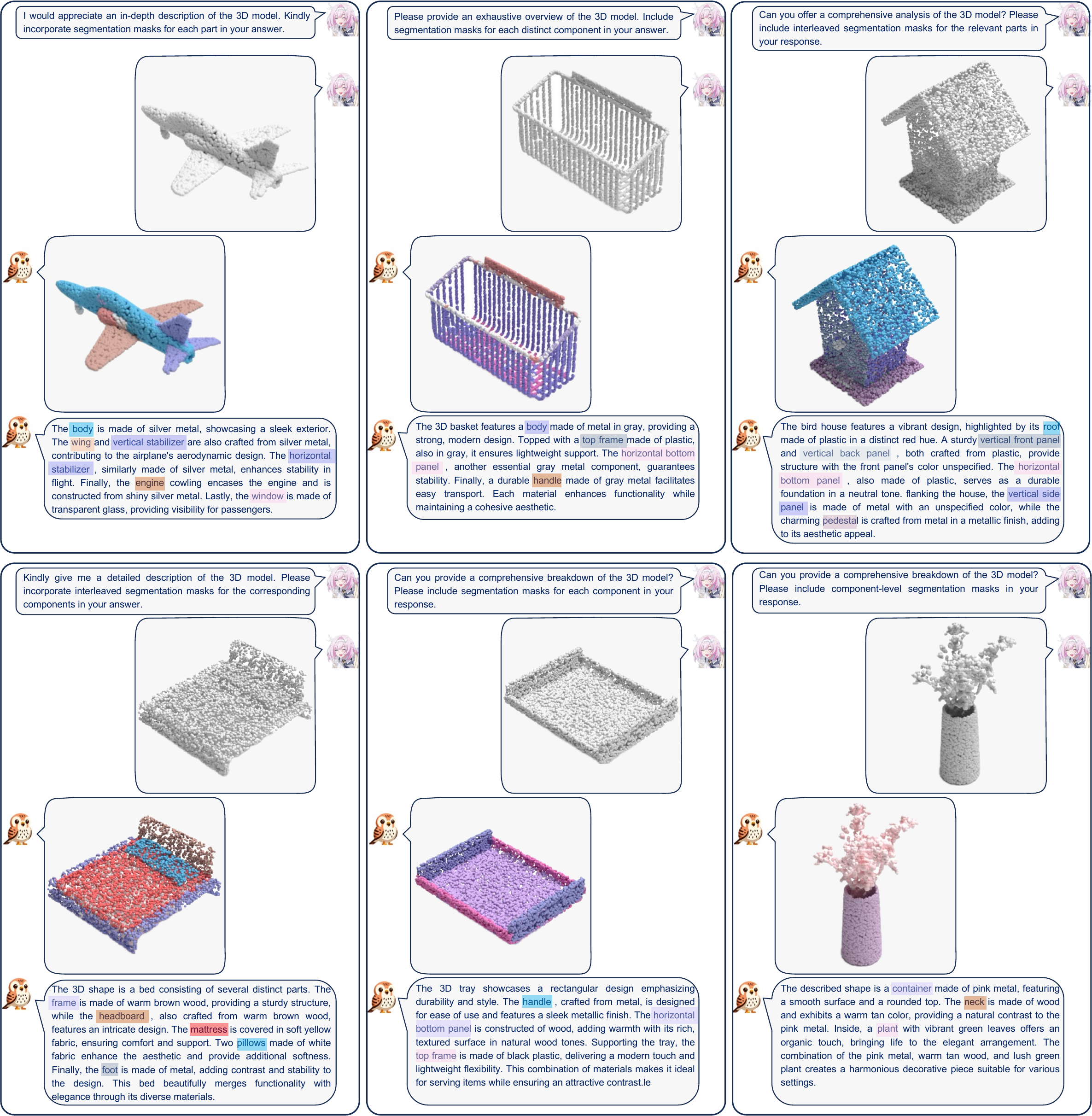}
    \caption{\textbf{Qualitative results of Kestrel on Part-Aware Point Grounded Description.}
    }
    \label{fig:supp_gd}
\end{figure*}

%% file: images/supp_single_part.tex
\begin{figure*}[t]
    \centering
    \includegraphics[width=1.0\linewidth]{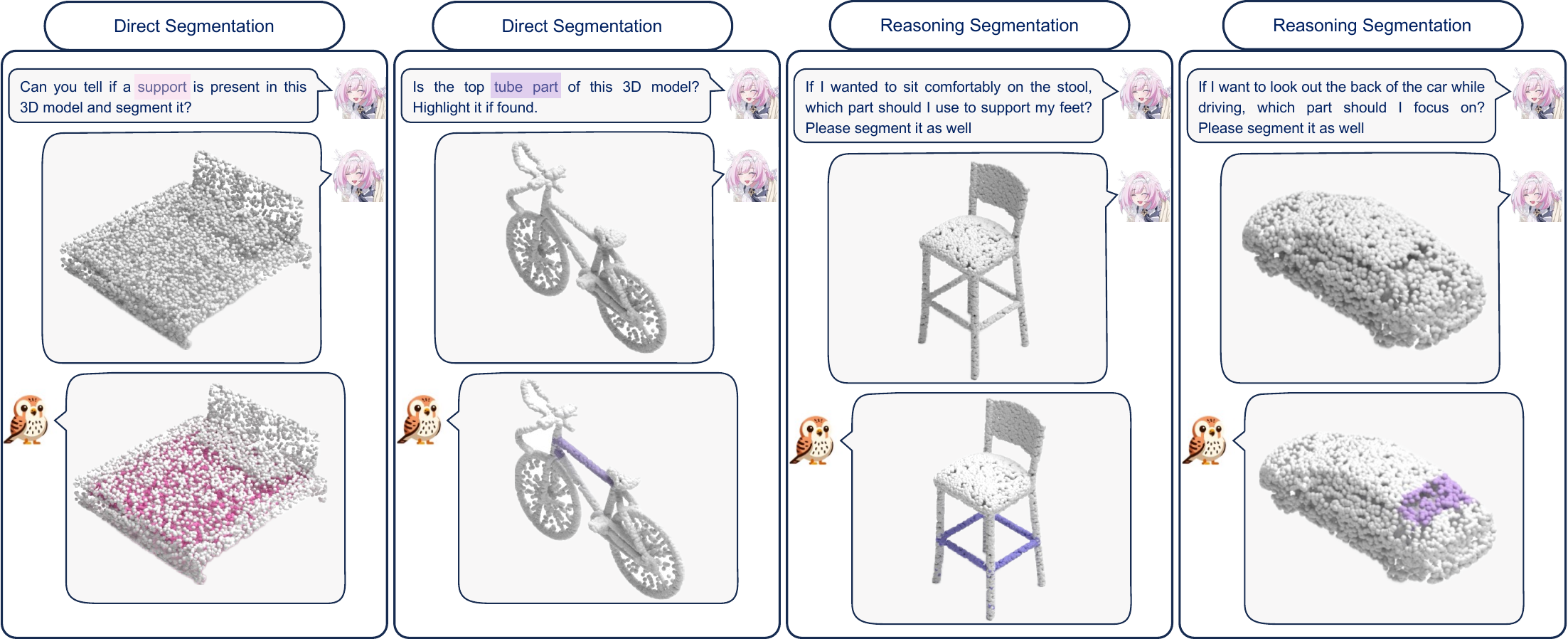}
    \caption{\textbf{Qualitative results of Kestrel on Single-Part Grounding.}
    }
    \label{fig:supp_sp}
\end{figure*}

%% file: sec/supp_baselines_arch.tex
\section{Mask3D Baseline}
\label{supp:baselines}

To extend the GLaMM model from 2D to 3D, we incorporate our multimodal large language model (MLLM), $\mathcal{F_{VL}}$, pretrained on PointLLM Objaverse~\cite{xu2023pointllm} and 3DCoMPaT-GRIN datasets, with the Mask3D segmentation model. Mask3D~\cite{Schult23ICRA} comprises a voxel-based ResUNet encoder, $\mathcal{E}_{voxel}$, and a segmentation decoder, $\mathcal{D}$, both pretrained on the ScanNet200~\cite{rozenberszki2022language} dataset. The Mask3D model is designed to process point clouds containing 8,192 RGB points with a voxel size of 0.01, employing intermediate voxel resolutions of 600, 1200, 2048, 4096, and 8192 within its ResUNet upsampling blocks.

During grounded caption generation for 3D shapes, $\mathcal{F_{VL}}$ outputs special [SEG] tokens, from which we extract the corresponding hidden states, $h_{seg}$. These hidden states are projected into the query embedding space of $\mathcal{D}$ and serve as positional queries for the segmentation decoder. The decoder then leverages these queries to predict part segmentation masks, aligning the textual grounding queries with the spatial representation of the 3D point cloud.

This process effectively combines the language understanding capabilities of $\mathcal{F_{VL}}$ with the spatial reasoning and segmentation strength of Mask3D, enabling robust 3D part-aware grounding. The process is described mathematically below:

\begin{equation}
    \mathbf{f}_{voxel} = \mathcal{E}_{voxel}(\mathbf{x}_{pc}),
\end{equation}
where $\mathbf{x}_{pc}$ is the input point cloud with 8,192 RGB points, and $\mathcal{E}_{voxel}$ represents the voxel-based ResUNet encoder that extracts voxel features $\mathbf{f}_{voxel}$.

\begin{equation}
    \mathbf{h}_{seg} = \mathcal{F_{VL}}(\mathbf{x}_{pc}, \mathbf{x}_{txt}),
\end{equation}
where $\mathcal{F_{VL}}$ is the MLLM that generates the grounded caption and produces hidden states $\mathbf{h}_{seg}$ corresponding to the [SEG] tokens.

\begin{equation}
    \mathbf{q}_{pos} = \mathcal{P}(\mathbf{h}_{seg}),
\end{equation}
where $\mathcal{P}$ is the projection layer that maps the hidden states $\mathbf{h}_{seg}$ into the query embedding space of the segmentation decoder $\mathcal{D}$.

\begin{equation}
    \mathbf{m}_{seg} = \mathcal{D}(\mathbf{f}_{voxel}, \mathbf{q}_{pos}),
\end{equation}
where $\mathcal{D}$ is the segmentation decoder, and $\mathbf{m}_{seg}$ represents the predicted part segmentation masks.

We set the max number of queries for Mask3D to 16

%% file: sec/supp_more_ablate.tex
\section{Additional Ablation Studies}
\label{supp:moreablation}

\begin{table}[h]
    \centering
    \resizebox{0.8\linewidth}{!}{%
        \begin{tabular}{cc|cccc}
        \toprule
        \multirow{2}{*}{\textbf{Point Enc.}} & \multirow{2}{*}{\textbf{Projector}} & \multicolumn{2}{c}{\textbf{Grounded Desc.}} & \textbf{D.S.} & \textbf{R.S.} \\
        ~ & ~ & 3D-CALC & mIoU & mIoU & mIoU \\
        \midrule
        PointBert & Q-Former & 43.80 & 72.24 & 67.40 & 64.9 \\
        PointBert & MLP      & 44.65 & 77.05 & 69.03 & 66.2 \\
        Uni3D     & Q-Former & 49.45 & 74.50 & 70.26 & 68.4 \\
        \rowcolor{Gray1}
        Uni3D (Final Choice) & MLP (Final Choice) & 50.10 & 86.70 & 80.70 & 71.8 \\
        \bottomrule
        \end{tabular}
    }
    \caption{\textbf{Ablation on different architecture designs}}
    \label{supp:arch_ablation}
\end{table}

\begin{table}[h]
    \centering
    \resizebox{0.75\linewidth}{!}{%
        \begin{tabular}{cc|c|c}
        \toprule
        \textbf{\# GD Samples} & \textbf{\# DS Samples} & \textbf{GD mIoU} & \textbf{DS mIoU} \\
        \midrule
        80K & 0 & 80.67 & 49.97 \\
        80K & 40K & 81.37 & \textbf{79.86}\\
        \rowcolor{Gray1}
        80K* & 8K* & \textbf{86.07} & 78.78 \\
        \bottomrule
        \end{tabular}
    }
    \caption{\textbf{Dataset Ablation.}}
    \label{supp:datamount}
\end{table}

\input{table/datasplit_ablation}

\paragraph{Model Architecture.} Tab.~\ref{supp:arch_ablation} presents an ablation study on different architectural choices for the point encoder and projector. The results indicate that both the choice of encoder and projector significantly influence the model's performance across all evaluation metrics. Using Uni3D as the point encoder consistently improves results over PointBert, regardless of the projector type. Similarly, MLP outperforms Q-Former as the projector, showing higher mIoU scores across Grounded Descriptions (GD), Direct Segmentation (DS), and Reasoning Segmentation (RS). The final chosen configuration, Uni3D with an MLP projector, achieves the best overall performance. This suggests that using an MLP to project the original point features into the LLM space produces tokens that closely match those used during upsampling in the segmentation decoder, leading to better results.

\paragraph{Dataset Amount.} Tab.~\ref{supp:datamount} Analyzes the impact of single-part data by varying the amount of Direct Segmentation (DS) data while keeping the Reasoning Segmentation (RS) data at zero as a control variable. Results show that adding a small amount of DS samples (8K) helps Kestrel learn part grounding. However, increasing DS to 40K raises training cost and slightly degrades performance on grounded descriptions. This suggests that a small amount of DS data offers a good trade-off for overall performance.

\paragraph{Dataset Distribution.} Tab.~\ref{reb:ablation2} explores how different training data subsets influence the model’s overall performance. Starting with only the grounded description subset yields moderate performance in each task. Adding the direct segmentation subset leads to a noticeable boost, particularly for its task. Finally, incorporating the reasoning segmentation subset achieves the best results, confirming that diverse training data covering grounded descriptions, direct, and reasoning-based instructions is essential for robust part-aware vision-language understanding. Together, these findings underscore the effectiveness of both progressive query refinement and comprehensive dataset coverage in enhancing language understanding and segmentation accuracy.

%% file: table/datasplit_ablation.tex
    
\begin{table}[b]
    \centering
    \resizebox{\linewidth}{!}{%
    
        \begin{tabular}{c|c|c|c}
        \toprule 
        \textbf{Training Data} & \textbf{Grounded Desc.} & \textbf{Direct Segmentation} & \textbf{Reasoning Segmentation}\\
        \midrule
        GD & 80.67 & 49.97 & 48.32\\
        GD+DS & 86.07 & 72.57 & 52.03\\
        \rowcolor{Gray1}
        GD+DS+RS & \textbf{86.70} & \textbf{80.70} & \textbf{71.8} \\
        \bottomrule
        \end{tabular}
    }
    \vspace{-7pt}
    \caption{\textbf{Further ablation on the dataset distribution.} GD: Grounded Dscription subset. DS: Direct Segmentation subset. RS: Reasoning Segmentation subset. }
    \label{reb:ablation2}
\end{table}

%% file: sec/supp_more_benchmark.tex
\section{PartNet-Mobility}
\label{supp:morebenchmarks}

\begin{table*}[ht]
    \centering
    \resizebox{\textwidth}{!}{%
        \begin{tabular}{c|cccccccccccccccc|c}
        \toprule
        \textbf{Model}& \textbf{Bottle}& \textbf{Chair} & \textbf{Display} & \textbf{Door} & \textbf{Knife} & \textbf{Lamp} & \textbf{Storage Furniture} & \textbf{Table} & \textbf{Camera} & \textbf{Cart} & \textbf{Dispenser} & \textbf{Kettle} &\textbf{Kitchen Pot} & \textbf{Oven} & \textbf{Suitcase} &\textbf{Toaster} & \textbf{Overall}\\
        \midrule
         PointNeXt &  68.4 & 91.8 & 89.4 & 43.8 & 58.7 & 64.9 & 68.5 & 52.1 & 33.2 & 36.3 & 26.0 & 45.1 & 57.0 & 37.8 & 13.5 & 8.3 & 50.2   \\
         PartSLIP & 83.4 & 85.3 & 84.8 & 40.8 &  65.2 & 63.9 & 66.0 & 53.6 & 58.3 & 88.1 & 73.7&  77.0&  69.6 & 73.5 & 70.4 & 60.0 & 59.4\\
         PARIS3D & 84.0 & 81.0 & 70.1 & 68.4 & 47.2 & 61.2 & 39.4 & 45.1 &29.3 & 71.7 & 40.1 & 59.3 & 78.8 & 59.1 & 61.6 & 24.9 & 57.6  \\
         PartSTAD & 83.6 & 85.1 & 82.3 & 61.4 & 63.8 & 68.3 & 59.5 & 47.7 & 64.3 & 85.0 & 73.7 & 84.2 & 73.5 & 71.8 & 68.2 & 58.6 & \textbf{65.0} \\
         Ours (zero-shot) & 67.6 & 58.0 & 57.9 & 56.0 & 56.5 & 58.3 & 32.4 & 40.1 & 35.1 & 59.1 & 43.6 & 67.5 &  72.4& 48.4 & 75.6 & 41.2 & 47.7\\
         Ours (few-shot) & 80.0 & 83.3 & 85.2 & 79.1 & 67.3 & 84.0 & 54.7 & 54.7 &  40.4 & 61.9 & 63.9 & 79.3 & 76.3 & 67.9 & 75.1 & 31.4 & \underline{\textit{63.9}}\\
        \bottomrule
        \end{tabular}
    }
    \caption{\textbf{Sample Category Results on PartNet-Mobility} Performance on selected categories and overall accuracy. All models are evaluated on the full set; only a subset is shown here.}
    \label{supp:partnet}
\end{table*}

Tab.~\ref{supp:partnet} presents a comparison of our method against prior baselines on a sample of categories from the PartNet-Mobility dataset, along with the overall accuracy. While PartSTAD achieves the highest overall score, it relies on training a separate model for each object category. In contrast, our method uses a single unified model across all categories. Notably, our few-shot setting outperforms or matches prior methods on several categories, including Lamp and Chair, and achieves strong overall performance. The results demonstrate the effectiveness of our approach in generalizing to diverse part segmentation tasks with limited supervision.

%% file: sec/supp_qualt.tex
\section{Qualitative Examples}
\label{supp:qualtitative}
Due to space constraints in the main paper, we present additional qualitative experiments in this section.
As shown in Fig.~\ref{fig:supp_gd}, Kestrel demonstrates its ability to provide comprehensive explanations of 3D objects, offering detailed, part-level descriptions for a given 3D object.
For each part-level phrase in the generated description, Kestrel predicts its corresponding position within the 3D space, represented by segmentation masks.
Fig.~\ref{fig:supp_sp} showcases the single-part segmentation grounding results. Kestrel demonstrates its ability to interpret part-aware instructions, understand user intent, and predict the corresponding position based on the given text input.

%% file: images/supp_gt.tex
\begin{figure*}[t]
    \centering
    \includegraphics[width=1.0\linewidth]{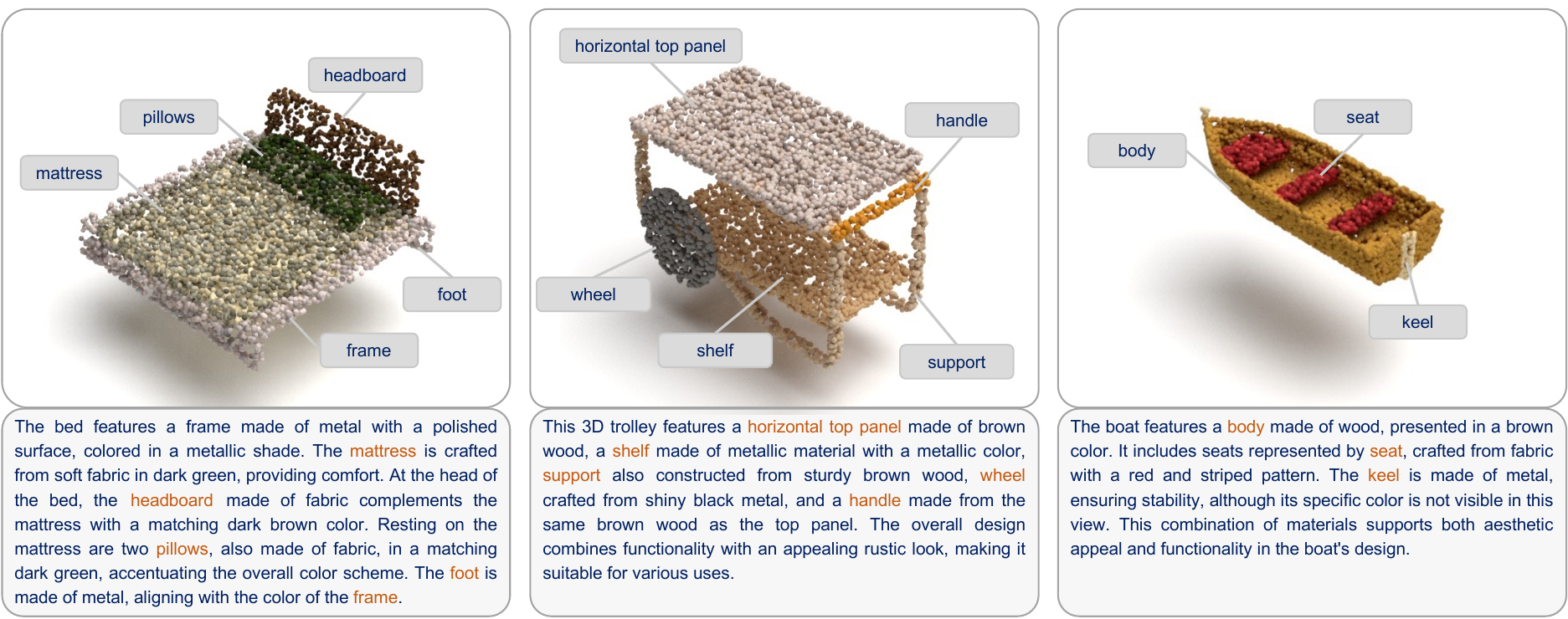}
    \caption{\textbf{Visualizations of collected \data~.}
    }
    \label{fig:supp_gt}
\end{figure*}

%% file: images/website.tex
\begin{figure*}[t]
    \centering
    \includegraphics[width=1.0\linewidth]{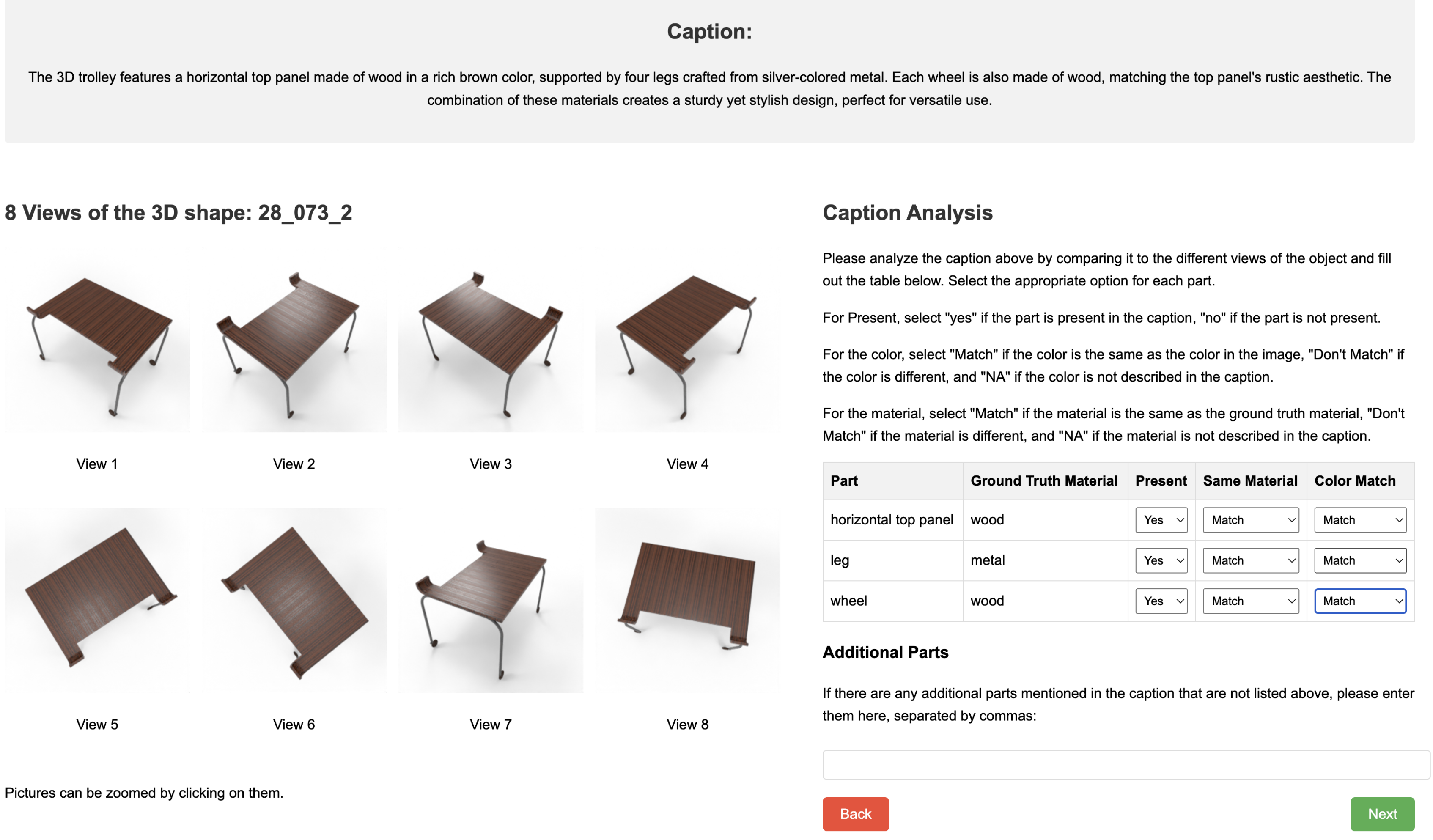}
    \caption{\textbf{Caption Validation Website.}
    The annotators are asked to compare the shape's caption with the ground truth parts and material and check that the part color described in the caption matches its color in the rendered images.
    }
    \label{fig:website}
\end{figure*}

%% file: sec/supp_data.tex
\section{Data Collection}
\label{supp:data}
\subsection{GPT4 collection}
In order to generate a detailed caption for each shape in 3DCoMPaT, we leverage the metadata of each shape to create part-material pair assignment text in the form of: "\textit{part\_name} made of \textit{material\_name}, $\cdots$". This metadata text is used in the GPT-4o prompt along with 8 views of the shape to caption it accurately in terms of the part, material, and color. a sample of the prompt can be seen in Tab.~\ref{tab:gpt_prompt}. For the reasoning segmentation task, we use the metadata to prompt GPT4o to create an indirect prompt and response each part. A sample of the prompt can be found in Tab.~\ref{tab:gpt_prompt2}

\subsection{Prompts}
A list of 30 predefined instructions is utilized to prompt the model to generate descriptions and ground segmentation masks, as detailed in Table~\ref{tab:instruction_list}. Additionally, another set of 15 predefined prompts, illustrated in Table~\ref{tab:prompts_single}, is used to evaluate direct segmentation task. To complement these instructions, a collection of 11 template responses, as shown in Table~\ref{tab:template_single}, guides the expected output format for the direct segmentation task. Each instruction and template was created with the assistance of GPT-4o to ensure diversity and relevance in the dataset.

The dataset creation process is divided into two distinct parts: 

\begin{enumerate}
    \item \textbf{Grounded Description Dataset}: This dataset is designed for multi-part grounding tasks, where each caption describes multiple parts of a shape along with their segmentation masks. To generate captions, we randomly assign an instruction from the 30 predefined instruction list to each sample and use GPT-4o to produce a descriptive and coherent output.
    
    \item \textbf{Reasoning Segmentation Dataset}: For this dataset, a single part from the shape is targeted for segmentation. A random instruction is selected from the single-part instruction set, and it is paired with a corresponding response template from the predefined list. This ensures consistency in both input and expected output formats.
\end{enumerate}

By leveraging these structured prompts and templates, we ensure that our dataset provides comprehensive coverage of both multi-part and single-part grounding tasks, effectively addressing the challenges of part-aware segmentation and language grounding in 3D models.

\subsection{Human Evaluation}

To evaluate our validation set of grounded descriptions, we conducted a human evaluation on the 6,770 grounded description samples. Annotators were provided with the interface shown in Fig.~\ref{fig:website} and were tasked with evaluating the following aspects:

\begin{enumerate}
    \item Whether the caption includes all the ground truth parts and their corresponding materials.
    \item Whether the part color described in the caption matches the true color visible in the rendered views.
    \item If the caption includes extra part names, we ask the annotator to mention them separated by a comma in order to correct the sample afterward.
\end{enumerate}

Based on the annotators' responses, we identify samples with inaccuracies and reran the GPT-4o pipeline on these samples. This iterative process ensured the creation of a fully accurate validation set, establishing a reliable benchmark for evaluating grounded descriptions.

\subsection{Examples}
Using the proposed dataset collection pipeline, we collect a total of 88,836 training samples and 6,770 validation samples for part-aware point-grounded description. Additionally, 677 validation samples are collected for single-part grounding.
Fig~\ref{fig:supp_gt} shows examples of the colored point clouds alongside their corresponding grounded descriptions.
As shown, the collected data effectively captures the various components of 3D objects, accurately representing each part-level component and its position.

%% file: table/gpt_prompt.tex
\begin{table*}[h]
\centering
\small
\begin{tabular}{p{0.9\textwidth}} 
\toprule
Given the following different views of the same 3D bench, caption the 3D shape by giving a description of the shape and its parts also describe the parts and their materials and add the exact color of each part from the provided part material assignment list: "seat is made of leather, seat\_frame is made of plastic, stretcher is made of plastic, leg is made of metal".
Make the caption short but comprehensive and descriptive. Your output must be without styling or line breaks and under 500 characters. You must mention the color of each part explicitly using their exact names from the list. Do not add any extra part names that are not on the list. Avoid describing the background or adding any unnecessary text or mentioning the words images, views, or objects. Do not use any words that shows you are not sure about the color. if the part name has \_ in it, Do not replace it with a space. If a part is not visible, do not mention that it is not found or not visible, instead mention the material description of the part. The caption should be coherent and descriptive.

The caption sentence is: \\

\bottomrule
\end{tabular}
\caption{\textbf{Sample GPT4o Prompt} An example of the prompt given to GPT4o }
\label{tab:gpt_prompt}
\end{table*}

%% file: table/gpt_prompt2.tex
\begin{table*}[h]
\centering
\small
\begin{tabular}{p{0.9\textwidth}} 
\toprule
I will be giving you a 3D object category along with part name and its material. I want you to generate a question prompt that inquires about the part's funtionality and usage and then provide the answer seperate by a new line. 
For example, if the object is a teapot, the part is the handle and the material is wood, the question should be \"If I wanted to hold the teapot, what part should I hold?\". You have to use the word 'part' in the question and mention the object name. Do not say the part name in the question and do not ask about its functionality since you mention the functionality in the question. The question must be descriptive of the part and uniquely identify the part in the object.
For the answer, you must use the same part name and the functionality of the part.
The object category is: car, the part name is: door, the material is: metal.
The question prompt is: \\

\bottomrule
\end{tabular}
\caption{\textbf{Sample GPT4o reasoning segmentation prompt} An example of the prompt used to create questions and answers }
\label{tab:gpt_prompt2}
\end{table*}

%% file: table/prompts_multi.tex
\begin{table*}[h]
\centering
\small
\begin{tabular}{p{0.9\textwidth}} 
\toprule
\begin{itemize}
\item Can you provide a comprehensive breakdown of the 3D model? Please include segmentation masks for each component in your response.
\item I would appreciate a thorough explanation of the 3D model. Kindly incorporate segmentation masks for the relevant parts in your answer.
\item Please offer a meticulous analysis of the 3D model. Include interleaved segmentation masks for the corresponding sections in your reply.
\item Could you give me an in-depth description of the 3D model? Please provide part-specific segmentation masks within your response.
\item I would like a detailed overview of the 3D model. Please include segmentation masks for each distinct element in your answer.
\item Kindly provide an extensive description of the 3D model. Please incorporate component-level segmentation masks in your explanation.
\item Can you offer a comprehensive analysis of the 3D model? Please include interleaved segmentation masks for the relevant parts in your response.
\item I request a thorough breakdown of the 3D model. Please provide segmentation masks for each part in your answer.
\item Please give me a detailed explanation of the 3D model. Include part-specific segmentation masks in your reply.
\item Could you provide an exhaustive description of the 3D model? Please include segmentation masks for each component in your response.
\item I would appreciate a meticulous analysis of the 3D model. Kindly incorporate interleaved segmentation masks for the corresponding sections in your answer.
\item Can you give me a comprehensive overview of the 3D model? Please provide segmentation masks for each distinct element in your explanation.
\item Please offer an in-depth breakdown of the 3D model. Include component-level segmentation masks within your reply.
\item I request a detailed description of the 3D model. Please incorporate part-specific segmentation masks in your response.
\item Kindly provide a thorough explanation of the 3D model. Please include segmentation masks for the relevant parts in your answer.
\item Could you give me an extensive analysis of the 3D model? Please provide interleaved segmentation masks for the corresponding components in your response.
\item I would like a comprehensive breakdown of the 3D model. Please include segmentation masks for each section in your reply.
\item Can you offer a meticulous description of the 3D model? Kindly incorporate part-level segmentation masks in your explanation.
\item Please provide an exhaustive overview of the 3D model. Include segmentation masks for each distinct component in your answer.
\item I request a detailed analysis of the 3D model. Please provide part-specific segmentation masks within your response.
\item Could you give me a thorough explanation of the 3D model? Please include interleaved segmentation masks for the relevant sections in your reply.
\item I would appreciate an in-depth description of the 3D model. Kindly incorporate segmentation masks for each part in your answer.
\item Can you provide a comprehensive breakdown of the 3D model? Please include component-level segmentation masks in your response.
\item Please offer an extensive analysis of the 3D model. Include segmentation masks for each distinct element in your explanation.
\item I request a meticulous overview of the 3D model. Please provide part-specific segmentation masks in your reply.
\item Kindly give me a detailed description of the 3D model. Please incorporate interleaved segmentation masks for the corresponding components in your answer.
\item Could you offer a thorough breakdown of the 3D model? Please include segmentation masks for each section in your response.
\item I would like an exhaustive explanation of the 3D model. Kindly provide part-level segmentation masks within your reply.
\item Can you give me a comprehensive analysis of the 3D model? Please include segmentation masks for the relevant parts in your answer.
\item Please provide a meticulous description of the 3D model. Include component-specific segmentation masks in your response.
\end{itemize} \\
\bottomrule
\end{tabular}
\caption{\textbf{Instruction list for grounding description task.} Each instruction is paired with a GPT-generated caption to guide the generation of part-specific segmentation masks for the 3D model.}
\label{tab:instruction_list}
\end{table*}

%% file: table/prompts_single.tex
\begin{table*}[h]
\centering
\small
\begin{tabular}{p{0.9\textwidth}} 
\toprule
\begin{itemize}
\item Does this 3D \textit{shape\_name} have a \textit{part\_name}? If yes, where is it located?
\item Is there a \textit{part\_name} in this 3D \textit{shape\_name}? Please segment it if it exists.
\item Can you tell if a \textit{part\_name} is present in this 3D \textit{shape\_name} and segment it?
\item Is a \textit{part\_name} included in this 3D \textit{shape\_name}? Please highlight its location.
\item Does this 3D \textit{shape\_name} contain a \textit{part\_name}? If so, please isolate it.
\item Please check if there is a \textit{part\_name} in the 3D \textit{shape\_name}, and segment it if present.
\item Is the \textit{part\_name} part of this 3D \textit{shape\_name}? Highlight it if found.
\item Is there a part described as \textit{part\_name} within this 3D \textit{shape\_name}? Segment it if present.
\item Can you confirm if the \textit{part\_name} exists in this 3D \textit{shape\_name} and segment it?
\item Does this 3D \textit{shape\_name} have a \textit{part\_name}? Show me where it is, if applicable.
\item Can you identify if the \textit{part\_name} is present in this 3D \textit{shape\_name} and segment it?
\item Please check if the area corresponding to the \textit{part\_name} is part of the 3D \textit{shape\_name}.
\item Can you verify the existence of a \textit{part\_name} and segment it within the 3D \textit{shape\_name}?
\item Is there a \textit{part\_name} in this \textit{shape\_name}? Segment it if found.
\item Check if the \textit{part\_name} is present in the 3D \textit{shape\_name}, and segment it if applicable.
\end{itemize}\\
\bottomrule
\end{tabular}
\caption{\textbf{Instruction list for direct segmentation task.} Each instruction is paired with a template answer. The \textit{part\_name} and \textit{shape\_name} words are replaced with the part and model names for each sample respectively.}
\label{tab:prompts_single}
\end{table*}

%% file: table/template_single.tex
\begin{table*}[h]
\centering
\small
\begin{tabular}{p{0.9\textwidth}} 
\toprule
\begin{itemize}
\item Yes, there is a \texttt{<p>}\textit{part\_name}\texttt{</p>[SEG]} part in the 3D \textit{shape\_name}.
\item Yes, a \texttt{<p>}\textit{part\_name}\texttt{</p>[SEG]} part is present in the 3D \textit{shape\_name}.
\item Confirmed, the 3D \textit{shape\_name} contains a \texttt{<p>}\textit{part\_name}\texttt{</p>[SEG]} part.
\item The \texttt{<p>}\textit{part\_name}\texttt{</p>[SEG]} part is found in this 3D \textit{shape\_name}.
\item Indeed, there is a \texttt{<p>}\textit{part\_name}\texttt{</p>[SEG]} part included in the 3D \textit{shape\_name}.
\item A \texttt{<p>}\textit{part\_name}\texttt{</p>[SEG]} part is present within the 3D \textit{shape\_name}.
\item Yes, the 3D \textit{shape\_name} includes a \texttt{<p>}\textit{part\_name}\texttt{</p>[SEG]} part.
\item There is a \texttt{<p>}\textit{part\_name}\texttt{</p>[SEG]} part in this 3D \textit{shape\_name}.
\item You can find a \texttt{<p>}\textit{part\_name}\texttt{</p>[SEG]} part in the 3D \textit{shape\_name}.
\item The 3D \textit{shape\_name} contains a \texttt{<p>}\textit{part\_name}\texttt{</p>[SEG]} part.
\item Yes, the 3D \textit{shape\_name} features a \texttt{<p>}\textit{part\_name}\texttt{</p>[SEG]} part 
\end{itemize} \\
\bottomrule
\end{tabular}
\caption{\textbf{Template list for direct segmentation task.} The \textit{part\_name} and \textit{shape\_name} words are replaced with the part and model names for each sample respectively.}
\label{tab:template_single}
\end{table*}